\documentclass[10pt,a4paper]{article}
\usepackage{amssymb,amsmath}
\usepackage{mathrsfs}
\usepackage{amsthm}
\usepackage{setspace}
\usepackage{color}
\usepackage{comment}
\usepackage{graphicx}
\usepackage{multirow}
\usepackage[all]{xy}
\usepackage{hyperref}
\usepackage{authblk}

\newcommand{\be}{\begin{eqnarray*}}
\newcommand{\ee}{\end{eqnarray*}}
\newcommand{\beqa}{\begin{eqnarray}}
\newcommand{\eeqa}{\end{eqnarray}}
\newcommand{\ba}{\begin{array}}
\newcommand{\ea}{\end{array}}

\newcommand{\mc}{\mathcal}

\newcommand{\mbb}{\mathbb}

\title{Low Impact Artificial Intelligences}

\author[1,2]{Stuart Armstrong%
  \thanks{Email: \texttt{stuart.armstrong@philosophy.ox.ac.uk}; Corresponding author}}
\affil[1]{The Future of Humanity Institute\\
Faculty of Philosophy\\
University of Oxford\\
Suite 1, Littlegate House\\
16/17 St Ebbe’s Street\\
Oxford OX1 1PT
UK}
\affil[2]{Machine Intelligence Research Institute\\
2030 Addison Street \#300\\
Berkeley, CA 94704}

\author[1]{Benjamin Levinstein%
  \thanks{Email: \texttt{balevinstein@gmail.com}}}

\date{2015}

\begin{document}
\maketitle

\begin{abstract}\noindent There are many goals for an AI that could become dangerous if the AI becomes superintelligent or otherwise powerful. Much work on the AI control problem has been focused on constructing AI goals that are safe even for such AIs. This paper looks at an alternative approach: defining a general concept of `low impact'. The aim is to ensure that a powerful AI which implements low impact will not modify the world extensively, even if it is given a simple or dangerous goal. The paper proposes various ways of defining and grounding low impact, and discusses methods for ensuring that the AI can still be allowed to have a (desired) impact despite the restriction. The end of the paper addresses known issues with this approach and avenues for future research.

\end{abstract}

\section{Introduction}

Imagine an artificial intelligence that has been given a goal such as `make paperclips', `filter spam in this account', or `cure this person's cancer'. If this AI is not very powerful, it is likely to attempt to achieve its goals in the ways we intend: improving industrial production, analysing and selectively filtering incoming messages, or looking for compounds able to differentially attack cancer cells. 

If the AI becomes very powerful, however, these goals all become problematic \cite{superI}. The goal `make paperclips' is perfectly compatible with a world in which the AI expands across the Earth, taking control of its resources to start an intense mass production of paperclips, while starting to launch colonisation projects for the other planets to use their resources for the same purposes, and so on. In fact, a naive version of the goal `make paperclips' mandates such actions. Similarly, `filter spam' is compatible with shutting down the internet entirely, and `cure this person's cancer' is compatible with killing her and destroying all the cells in her body.

There are several proposed approaches to combat this issue. The most standard is to add something to the goal, fleshing it out so that it includes safety components (`and don't kill anyone, or inadvertently cause their deaths, or...'). As the AI's power increases, its potential influence over the world increases as well, and the safety components need to be fleshed out ever more (`...and don't imprison people, or cause a loss of happiness or perceived liberty or free will, or...'). The `Friendly AI' approach aims roughly to specify these safety components in as much specific detail as possible \cite{eliezer_AI_pm_risk}. Other approaches aim to instil the these components via implicit or explicit learning and feedback \cite{learnValue,mot:val}.

This paper takes a different tack. Instead of specifying the safety components, it aims to ensure AI has a low impact on the world. Given this low impact, many otherwise unsafe goals become safe even with a very powerful AI. Such an AI would manufacture a few more paperclips/filter a few more messages/kill a few cancer cells, but would otherwise not take any disruptive action.

The first  challenge is, of course, to actually define low impact. Any action (or inaction) has repercussions that percolate through the future light-cone, changing things subtly but irreversibly. It is hard to capture the intuitive human idea of `a small change'. 

There are a few intuitive ways in which an action can have a low impact, though, which we examine in some depth in Section \ref{sec:DifferentMeasures}.  For example, if we can describe the universe in terms of a huge number of disparate  but well-chosen variables and the action has little impact on their values, then it was not of high impact. We can also assess whether knowing the action is particularly `important' in terms of predicting the future, or whether we can see if the actions are likely to be detectable at a later date. If the action is such that any difference to the universe is lost in entropy or absorbed into a chaotic and unpredictable process, it certainly has a low impact. Finally, we can also abstractly compare the features of probability distributions of future worlds given the action or not. 

The second challenge, tackled in Section \ref{sec:HighImpact}, is to figure out how to ensure that the AI's impact is not too low -- that we can still get useful work out of the AI, without risking a larger or negative impact.  Although low impact seems to preclude any action of significance on the part of the AI, there are a number of ways around this limitation. Unlike the bad AI impacts that we are trying to prevent, we can often have a much clearer idea of the positive impact we are attempting to accomplish. This allows us to construct various specific setups and targeted loopholes to allow specific high impact from otherwise low impact AIs.

The final, brief, section looks at some of the problems and unresolved issues with the current setup -- and hence the directions for future research.

\section{The General Framework}
\subsection{The penalty function}

Although determining what exactly counts as `impact' will be a thorny issue, we can nonetheless characterise the approach abstractly. The basic idea is that the AI has some active goal, such as \emph{cure cancer} or \emph{filter spam}, but it wants to pursue this goal without changing the world in any important way. We can then describe its utility function as follows:
\begin{align} \label{penalty:equation}
	U = u - \mu R.
\end{align}
The function $u$ is a standard utility function that gives the AI its active goal. The function $R$ is the penalty function, penalising the AI for having a large impact. The number $\mu$ is some scaling factor, setting the importance of low impact relative to the AI's active goal $u$.

In order to prevent the AI accepting a large $R$ penalty in exchange for a large $u$ gain, we will want to define a bounded $u$, such that performance close to the maximum bound is not too difficult to obtain. There is no such bound on $R$, of course: the more impact the AI has, the more it gets penalised.\footnote{%
Note that equation \eqref{penalty:equation}, if it works as intended, makes the AI relatively transparent to humans. The active goal $u$ can be simple, while the intent of low impact is clear. 
}

\subsection{Defining the alternatives}



To define low impact, we first will need a baseline for comparison. What is a low impact, as opposed to a non-low one? The most natural alternative, and the one we'll use, is the world in which the AI is never successfully turned on in the first place, or, to be more precise: some prior probability distribution $P$ over the set of worlds $\mc{W}$ conditional on the AI not having been turned on. An AI that was never turned on is assumed to have very low impact; the behaviour of the active AI is compared with this baseline.

For such a distribution to make sense, we'll assume the turning on of the AI does not occur with probability $1$. For instance, we can make it depend on a signal going through a specific wire, which goes through some noisy environment -- say, an unstable gas -- that has a tiny chance of interrupting the signal. If the signal is interrupted, the AI is not turned on.

Designate the event of the signal passing through the wire as $X$ (which then turns on the AI), and the signal being blocked as $\neg X$ (which then leaves the AI off). We then want the AI to have a low impact on the world, with a low impact being informally defined as the world, conditional on $X$, being `not too dissimilar' from the world, conditional on $\neg X$. The purpose of this paper is to make this definition rigorous\footnote{%
Instead of comparing the world conditional on $X$ with the world conditional on $\neg X$, it may be desirable to consider various kinds of subjunctive or counterfactual suppositions instead if we're especially interested in, say, $X$'s expected causal consequences and not in its evidential impact. For the sake of simplicity, however, we'll restrict attention to the standard notion of conditional probability instead in this report.
}.

Since conditional probability does not depend on the probability of the event being conditioned on, we would want the probability of $X$ to be almost $1$ -- the AI is almost certain to be turned on. This is consistent with the fact that, having built an AI, the designers would certainly want to make use of it\footnote{%
There are many subtleties here, the main ones of being that if $\neg X$ happens, humans are likely to question their probability estimates, and will likely just try turning it on again. Humans concluding that something is wrong can be dealt with via methods similar to \url{http://lesswrong.com/r/discussion/lw/lxl/humans_get_different_counterfactuals/}. Preventing the AI from taking into account future human attempts to turn it on again could be addressed by adapting the methods proposed in \url{http://lesswrong.com/r/discussion/lw/luy/acaucal_trade_barriers/}. But a detailed discussion of this is beyond the scope of this current report.
}.

\subsection{Boxing the AI}

One useful precaution is to `box' the AI, confining it to a sealed environment somewhat isolated from the outside world. Unlike other boxing suggestions \cite{oracleAI, superI, leakproof_sing}, the aim here is not to prevent the AI from having an impact by confining it. Rather, it is to minimise the difference between worlds with $X$ and $\neg X$. The box is not to prevent the AI from acting (the definition of low impact will ensure that), but rather to prevent us from seeing that the AI is active. Thus any visible impact of the AI will come only from via the AI's decisions, rather than from it's mere existence.

\section{Different measures of low impact}\label{sec:DifferentMeasures}
\subsection{The fundamental challenge}

The fundamental challenge of developing the low impact approach is that, from a physical point of view, everything done or not done has an impact (a probabilistic one, at least), across all possible futures. Therefore the world, given $X$, must be fundamentally different from the world, given $\neg X$, simply because of the difference between $X$ and $\neg X$. Particles will be arranged differently in these different scenarios, and this effect will compound the further we get from the event.

What we're really interested in `no important impact'. It matters little if the atoms that define the US president are a little to the left of where  they would be otherwise. It matters a lot that the AI doesn't escape from its box, eliminate all humans and optimise the universe in the service of goal $u$.

A second component of this challenge is that the AI may end up creating an importantly different world, simply because some processes  such as the weather are chaotic and $X$ versus $\neg X$ change the initial conditions. Yet in these cases, its impact should nonetheless count as low. Suppose, for instance, the upcoming election would be very close. The AI makes breakfast on election day for a capricious and unpredictable voter who ends up voting for Alice instead of Bob in part because he had apricots instead of biscuits to eat. As a result, Alice is elected instead of Bob. Although the effect of the breakfast decision is large, it ought not be considered `high impact', since if an election was this close, it could be swung by all sorts of minor effects. We'll therefore investigate probabilistic approaches: what sorts of changes can be predicted \emph{ex ante} if $X$ is true?

\subsection{Coarse graining: Twenty billion questions}\label{coarse:g}

One way to solve the fundamental challenge is first to find a way of `coarse-graining' the set of worlds. That is, we partition the set of worlds into small cells, and any two elements of the same cell count as equivalent for our purposes. Generally these cells will be determined by the values of certain variables or characteristics. We can then measure impact in terms of the AI's expected effects over this partition. Thus if $A$ is any element of this partition, we typically expect the probabilities $P(A|X)$ and $P(A|\neg X)$ to be close.\footnote{%
Note that the AI's limitations make this approach easier. If the AI has difficulty predicting $A$ precisely (maybe because $A$ is chaotic or because of the fundamental entropy limitations on information), then those two probabilities are likely to be closer than otherwise. This helps to distinguish small inadvertent impacts the AI might have from the larger deliberate ones we want to avoid.
}%



The purpose of coarse graining is to define the world in sufficient detail that the AI cannot have a large impact without disrupting most of those details.
The characteristics used must be as broad and as diverse as possible, making it impossible for the AI to grain great power without disrupting some of them.\footnote{%
It's important to note that an extremely powerful AI could potentially gain control of the world and set the characteristics to whatever values it wants. Therefore, the purpose of the coarse graining approach is to catch the AI  before it initiates sweeping changes to the world.} %
For instance, we could use the air pressure in Dhaka, the average night-time luminosity at the South Pole, the rotational speed of Io, and the closing numbers of the Shanghai stock exchange. To be confident that we can sufficiently constrain a super-intelligent AI, we'll need millions if not billions of these variables, separating the universes into uncountable numbers of different cells\footnote{%
Though not uncountable in the sense of an uncountable infinity, the number of cells is exponential in the number of characteristics/variables, and is thus easily larger than the number of particles in the reachable universe. Therefore, as we add variables, the difficulty for the AI to outsmart us and find an unintended high-impact way of maximising $u$ should grow fairly rapidly, since all constraints must be met simultaneously.
}.

Of course, we have to take care in deciding which characteristics to use. This procedure doesn't work if some variables are too directly connected with $X$ or $\neg X$. The electricity consumption of the AI, for example, cannot be a variable, but the electricity consumption of the entire county could be if the AI's usage is likely to be lost in the noise. But we can hope that for ``large scale'' issues, that the universe is ``roughly'' unchanged given $X$ and $\neg X$. 

For any world $w$, we can define a world vector $V_w$ which is the values taken by the billions of chosen variables\footnote{%
Continuous variables will need a sensible binning.
}. We can then make use of vectors to coarse grain the space of worlds, defining an equivalence relation:
\begin{align*}
	v \cong w \textrm{ iff } V_v = V_w
\end{align*}
The cells are the equivalence classes of this relation. This allows us to make statements like `the probability of $w$ is equal whether or not the AI was turned on'. Without the course graining, $P(w|X)=0$ or $P(w|\neg X) = 0$, depending on whether or not $w$ contained the AI. But with the coarse graining, the statement becomes:
\begin{align*}
	P(V_w | X) = P(V_w | \neg X)
\end{align*}
And, as long as those variable values are possible given $X$ and given $\neg X$, the above formula makes sense.

\begin{figure}[h!tb]
	\centering
	\includegraphics[width=0.7\textwidth]{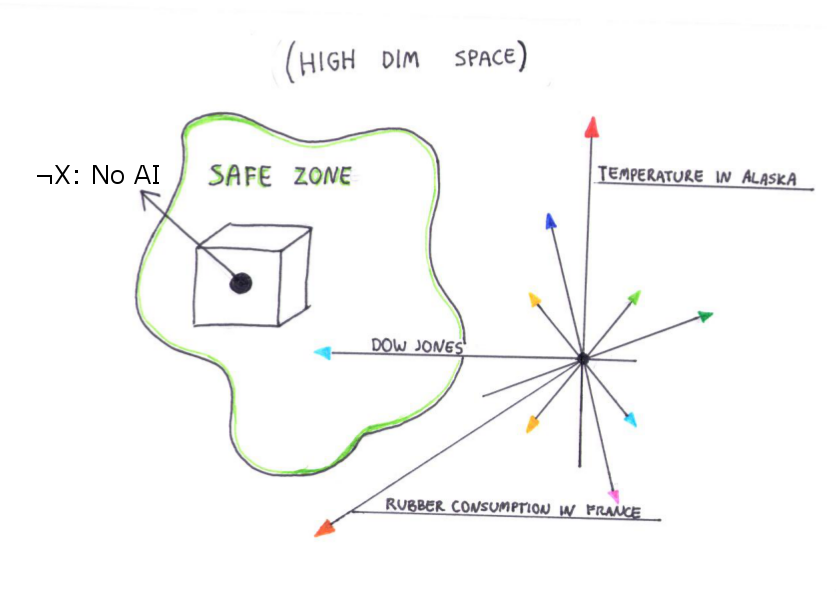}
	\caption{The many variables define a `box' in high dimensional space, which includes the $\neg X$ point where there is no AI. The AI must maneuver within this box, which hopefully should reside wholly within the safe zone.}
	\label{safe:box}
\end{figure}

Then we can formulate low impact as some sort of measure of the difference between the expected worlds given $X$ and $\neg X$. The $l_{\infty}$ norm, for instance, could work:
\begin{align}\label{coarse:eq}
	R = \max_{w} |P(V_w | X) - P(V_w | \neg X)|
\end{align}
The `box' defined by this norm is illustrated in figure \ref{safe:box}. `Softer' versions of this maximum norm could work as well.

An important feature of the approach is that we have no need to measure these billion variables precisely. They are simply used to define the coarse graining, with the AI then trying to get its best estimate of $R$. Nor do they need to perfectly define the variables: the closing numbers of the Shanghai stock exchange could be defined as a specific output along a specific wire. In worlds where the stock exchange didn't exist, that input would correspond to something else entirely -- maybe blank signals, maybe noise. But this shouldn't matter, as it still coarse grains `worlds with stock exchanges emitting along this wire, given $X$' together in the same cell with `worlds with stock exchanges emitting along this wire, given $\neg X$' and similarly for worlds without\footnote{%
Note that this is in contrast with more traditional AI approaches, where the terms have to be correctly defined \cite{eliezer_AI_pm_risk} or grounded \cite{symbol_ground}. The purpose is not to have the meaning of each variable defined precisely, but to capture so many variables that they limit the AI's ability to have large impacts.
}.

\subsection{The importance of knowing \texorpdfstring{$X$}{X}}


A related option is to measure specifically the informational value of the AI on things we might care about. The intuitive idea is that if the AI is high impact, then knowing whether or not the AI was turned on tells you a lot more about the future of the world than another generic piece of information. We can then set the penalty $R$ to measure the importance of $X/\neg X$ with respect to other pieces of evidence. If the AI has truly low impact, then if presented with a large number of facts about the world, $X$ will not stand out as being more informative than other facts.

How would we measure importance? One approach is to look at a large set $\mc{U}$ of different utility functions, and check that knowing $X$ makes little expected difference for any of them\footnote{%
One method for choosing $\mc{U}$ is to have the AI or some other superintelligence select a class of utility functions that it thinks is likely to contain the set of actual or potential \emph{human} utility functions, or some near enough approximation. That is, we want $\mc{U}$ more or less to contain representations of everything humans may really care about. If the AI does not, in expectation, affect the elements of $\mc{U}$ that much, then it counts as low impact in the desired sense of the term. Just as in the coarse graining case, we'd want to add a variety of other utilities to $\mc{U}$, in case there are vulnerabilities that have not occurred to us. We of course will need to exclude utility functions that care directly about whether $X$ for this approach to work.
}.

When measuring importance, we could either check the expected difference of knowing $X$, or the expected importance of other facts, conditional on knowing $X$. More formally, let $\mc{F}$ be a large set of facts compatible with both $X$ and $\neg X$, and define:
\begin{align*}
	R = \max_{u' \in\mc{U},\mc{S}\subset\mc{F}} \left|\mbb{E}(u'|\mc{S}\cup X) - \mbb{E}(u'|\mc{S}\cup \neg X)\right|
\end{align*}

\subsection{Undetectable means unimpactful?}\label{undetect}

This sections looks at another way of defining low impact: undetectability. If, during the 19\textsuperscript{th} century, there was an inhabitant of London, and there now remains no record whatsoever of their existence, it is likely that they had a very low impact. Presume that we are given a partial description of them, and asked, from the point of view of today and all the evidence we have, whether they existed or not. If we really cannot tell either way, this seems a definitional example of low impact: their potential impact is literally lost in the noise of the world\footnote{%
Of course, some care must still be taken when we decide how to describe this person. For instance, we can't tell now whether there really was somebody called `Jack the Ripper' who was a serial killer and had exactly 100,000 hairs on his head at midnight on 1 January 1888. The more formal development of this approach below will eliminate this issue.
}.

How could this be formalised? Suppose that the AI uses a probability estimator $P'$. This estimator is bounded by practical concerns, as any real-world agent can only have bounded rationality. Its unbounded equivalent is $P$. And these estimators have some level of reflection \cite{reflection}, so $P'$ can express some estimates over the outcome of $P'$ (self-reflection) and $P$ (modelling of a superior rational agent). Let $b$ designate background information about the world (prior to $X$), and let $\mc{G}$ designate the set of events in a slice $S$ of event $X/\neg X$'s future light-cone -- maybe the slice $50$ (time-like) years in the future.

Then what we would want is that $P(g|X,b) = P(g|\neg X, b)$ for $g\in\mc{G}$ -- that the likely future is identical whether or not the AI is turned on or not (as assessed by the unbounded version of the agent, with access to all of $S$). If we designate the expectation estimator derived from $P'$ as $\mbb{E}'$, this reduces to wanting:
\begin{align*}
	R = \mbb{E}'( | P(g|X,b) - P(g|\neg X,b)| ) = 0.
\end{align*}
This expectation is both over $P$ (the AI's estimate of its unbounded estimator) and over the likely $g \in \mc{G}$ (the likely future; this also depends on the AI's policy choices).

This works if the impact of the AI is really lost through entropy, such that even $P$ cannot detect it. However, an unbounded agent with full access to the slice $\mc{G}$, might be able to always detect the presence of an AI, no matter how low impact it attempts to have.

In that situation, we can consider the change in $P$ as it's fed information from $\mc{G}$, one bit at a time. When $P$ has all of $\mc{G}$, it will detect the presence of the AI, but before that point, it will have doubts. An AI that optimises the world is likely to leave many traces all over $\mc{G}$; one that changes little will leave corresponding few choices. For $0\leq\rho\leq 1$, designate by $g^{\rho}$ a random sampling of $\mc{G}$ where we have randomly chosen a proportion $\rho$ of all the data in $\mc{G}$. We can designate a new measure of R:
\begin{align}\label{undetect:equation}
	R = \min_{\rho} \left\{\mbb{E}'\left(\frac{P(g^{\rho}|X,b)}{P(g^{\rho}|\neg X,b)}\right) >10 \textrm{, or } \mbb{E}'\left(\frac{P(g^{\rho}|\neg X,b)}{P(g^{\rho}|X,b)}\right) >10 \right\}
\end{align}

Here $R$ is a measure of how much information $P$ is likely to need before detecting the difference between $X$ and $\neg X$ (`difference' being somewhat arbitrarily defined as the expected ratio of probabilities, or inverse ratios, being higher than $10$).

\subsection{Comparing distributions abstractly}\label{comp:abs}

To avoid some of the issues and choices needed for the options discussed so far, we can attempt to compare the two conditional probability distributions $P(-|X)$ and $P(-|\neg X)$ abstractly. The intuition behind this is that if we have a low impact AI or no AI at all, the probability distribution over future worlds is likely spread far and wide across many different options, but if there is a powerful unconstrained AI, the future is likely to be highly optimised causing $P(-|X)$ to take on a different shape. We can therefore compare the conditional distributions by looking at various descriptors they might have -- means, variances, peaks, and so on. Such descriptors depend on the underlying metric on the set of worlds $\mc{W}$, and so ideally these descriptors will be very similar for a wide variety of natural metrics we could choose.

The weakness of this approach is that we have an extremely reduced understanding of what is going on in the comparison, so we don't know if the AI can produce a distribution $P(\mc{W}|X)$ that is formally similar to $P(\mc{W}|\neg X)$, but in practice is a very concentrated, $u$-optimised world. However, perhaps with a sufficiently robust statistical characterisation of the relevant distributions, this won't be an issue.

\subsubsection{Generalised Cross-Entropy}

One natural abstract way of measuring the expected impact of $X$ is to compare the divergence between $P(\mc{W}|X)$ in terms of $P(\mc{W}|\neg X)$. If the two distributions are relatively close, then $X$ likely does not have an especially large impact.

Unfortunately, it's not obvious what particular measure of divergence we ought to use. Kullback-Leibler divergence---the standard measure---won't work in this case. Let $P_X=P(\mc{W}|X)$ and $P_{\neg X}=P(\mc{W}|\neg X)$. $P_X(X)=1$ and $P_{\neg X}(X)=0$, so $D_{\text{KL}}(P_{\neg X}||P_X)=\infty$.

There are, however, other measures of generalised entropy and divergence that are bounded and may be able to do the job. Bounded Bregman-divergences, for instance, are often used to quantify the amount of generalised information needed to move from one probability function to another.\footnote{%
The precise details of generalised measures of entropy and Bregman divergences and their relationship to information theory are involved and not worth expounding in detail here. For extended discussions, see \cite{GneitingandRaftery07} and \cite{GrunwaldandDawid04}.
}
Whether such an approach will work for our purposes remains to be seen.



%

\section{High impact from low impact}\label{sec:HighImpact}

All the preceding methods aim to reduce the impact of the AI. Of course, we don't actually want a low impact overall -- we want a low negative impact. The problem is that we cannot successfully define ahead of time what these negative impacts are.

So how can we ensure that we actually get some level of \emph{positive} impact from using such AIs?

\subsection{Calibrating the penalty function}

The most obvious option is to `tune the dial' in equation (\ref{penalty:equation}) by changing the value of $\mu$. We can start with a very large $\mu$ that ensures no impact at all -- the AI will do nothing. We can then gradually reduce $\mu$ until we get an action that actually increases $u$.

\begin{figure}[h!tb]
	\centering
	\includegraphics[width=0.7\textwidth]{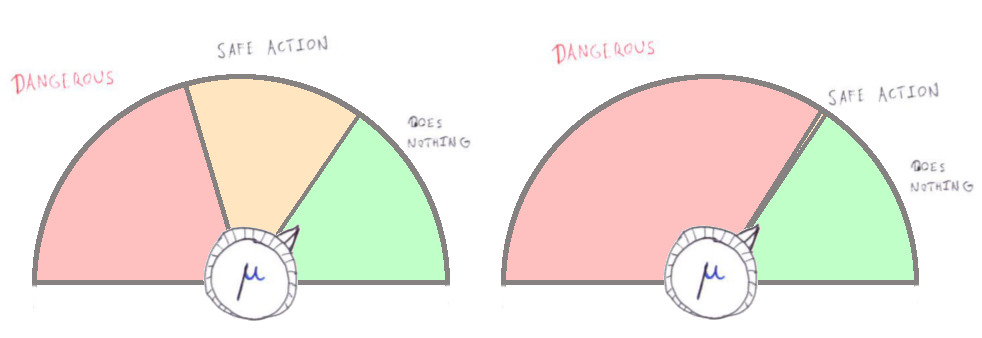}
	\caption{Though it may intuitively feel there is a large zone between the AI doing nothing and having a dangerous behaviour (dial on the left) this need not be the case (dial on the right).}
	\label{dial}
\end{figure}

This does not seem especially safe, however. The first issue is that we have little understanding of the correct value for $\mu$, so little understanding of the correct rate to reduce $\mu$ at. It is conceivable that we spend a million steps reducing $\mu$ through the `do nothing' range, and that the next step moves over the `safe increase of $u$', straight to the `dangerous impact' area. In other words, there may be a precipitous jump from the level at which $\mu R$ dominates $u$, to the level at which $u$ becomes sufficiently unconstrained by $\mu R$ to lead to dangerous behaviour. See figure \ref{dial} for illustration.

The central failure, however, is that in many cases it is not clear that low impact is compatible with \emph{any} increase in $u$. 
In particular, when it's clear that the AI has done something, low impact might be impossible. Even the simple fact that the AI had done anything might get reported, passed on, commented upon. It might affect the whole future development of AI, economic policy, philosophy, and so on. This might disrupt any effect of low impact (e.g.~any action the AI takes might have an impact the AI can predict), meaning that there is \emph{no} safe range for $\mu$: the AI must either do nothing, or have a large impact.

However, though we cannot successfully define the negative impacts of the AI we wish to avoid, we are on much firmer grounds when defining the positive aim we are looking for. This suggests other ways of producing higher impact: by specifically allowing what we want to allow.

\subsection{Unsafe output channel}\label{output:cond}
The easiest way to allow for higher impact, is to specifically exclude the AI's output from the low impact requirements. For instance, assume the AI is going to send out message $O$. To ensure that $P(O|\neg X) \neq 0$, we set up an alternative system that will produce a random message. 

Then we exclude the contents of $O$ from the low impact considerations. For instance, we can modify the coarse-graining approaching developed above by replacing equation (\ref{coarse:eq}) with $R = \max_{w} |P(V_w | X, O) - P(V_w | \neg X, O)|$. Alternatively, we can abstractly compare $P(\mc{W}|X, O)$ with $P(\mc{W}|\neg X, O)$, or replace equation (\ref{undetect:equation}) with:
\begin{align*}
	R = \min_{\rho} \left\{\mbb{E}'\left(\frac{P(g^{\rho}|X,b,O)}{P(g^{\rho}|\neg X,b,O)}\right) >10 \textrm{, or } \mbb{E}'\left(\frac{P(g^{\rho}|\neg X,b,O)}{P(g^{\rho}|X,b,O)}\right) >10 \right\}.
\end{align*}

\begin{figure}[h!tb]
	\centering
	\includegraphics[width=0.7\textwidth]{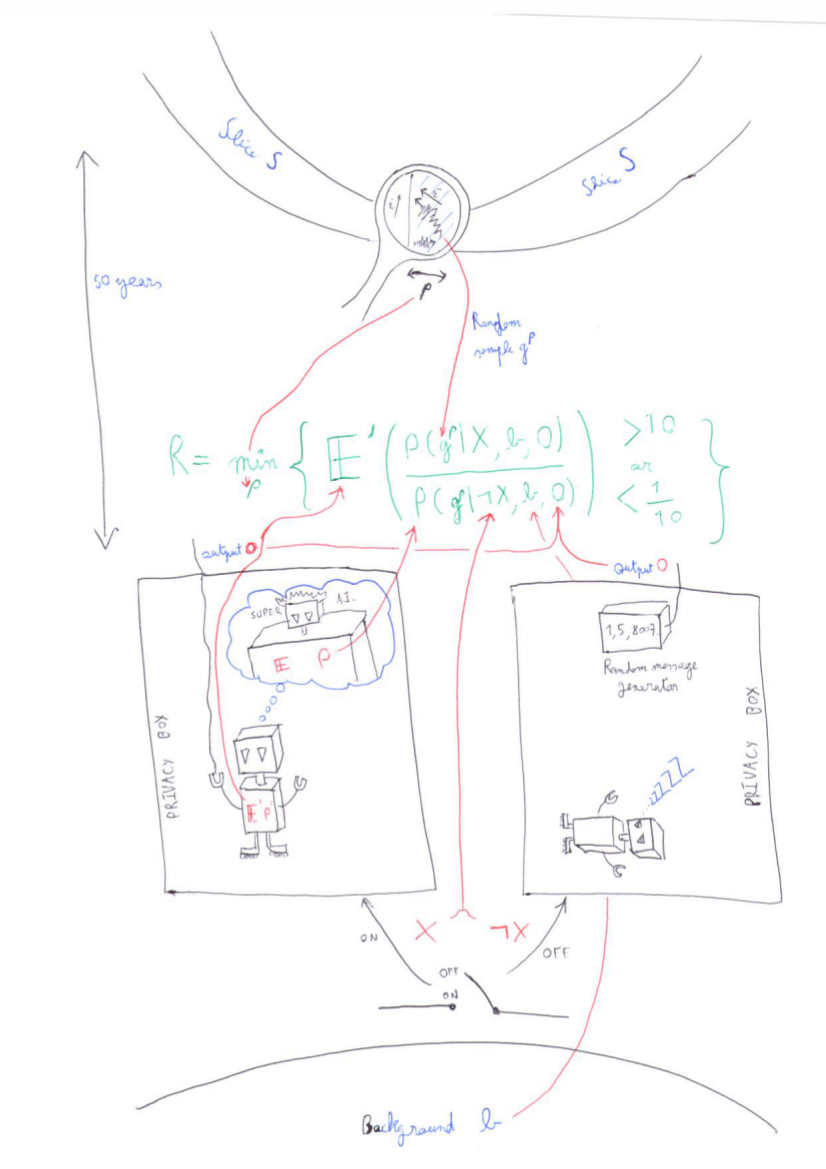}
	\caption{The various elements that go into defining the penalty function for a low impact AI aiming to be undetectable but allowed an unsafe output channel.}
	\label{no-info-output}
\end{figure}

All of these conditionings reflect the fact that though a coherent $O$ is likely to present very strong evidence for $X$ over $\neg X$, we can control for any \emph{further} evidence -- such as the evidence provided by an AI expanding and optimising the world. A diagram of the setup for the `undetectable' approach of section \ref{undetect} is given in figure \ref{no-info-output}.

This type of low impact keeps the AI from having an impact, outside of the content of the message $O$. \emph{It does not make the message $O$ itself safe} -- $O$ is excluded from any low impact requirements. Therefore, the use of this method is to block other routes for AI influence, and confine the risk to the output itself, which can still be very dangerous \cite{superI}.

\subsection{WBE, chatbots, and conditioning on results}\label{res:con}

As we just observed, one of the problems of low impact is the content of any message the AI sends out. Indeed, unless we condition away that message as in the previous subsection, the very existence of a coherent message is a potentially massive indication of impact: it means the AI was turned on and did something.

So one way of dealing with this is to have something else in the box that could output a coherent message. Maybe a real human\footnote{About to die, so never able to talk about this experience, or sworn to eternal silence?}, a whole brain emulation \cite{WBERM}, another less intelligent artificial agent, a chatbot, or maybe a random selection of internet messages. If the AI is not turned on -- if $\neg X$ -- then the other agent will output some message on the subject. So the emission of a coherent message \emph{per se} is no longer surprising, nor direct evidence for the AI being turned on.

In that situation, the low impact setting now motivates the AI to do a perfect imitation of a chatbot/human/WBE -- and nothing more. But we can get around this restriction by conditioning on something other than the actual content of the message.

Suppose that we want the AI to make us rich -- maybe with stock market advice. Then, in a year's time, if we have, say, multiplied our fortune by ten,  we will announce ``the AI's advice was good''. Call this announcement $A$. Then we use $A$ to condition on as in Section \ref{output:cond}, instead of $O$.

What does this mean in practice? It means that the AI is committed to low impact, \emph{given that we multiply our fortune by ten}. So, the world in which we get a random lucky message $O$ and then get rich must resemble the world in which the AI gives us advice $O$ and we get rich\footnote{%
Note that we condition on the announcement $A$ (which is easy to define) not on the concept of getting rich (or whatever else our goal is) as that might be harder. This would be very dangerous if we told the AI `ensure $A$ happens', as $A$ is only loosely connected with our true goal. But here we are using $A$ as a sufficient measure of success in a world in which the AI doesn't exist. In that world, it is a safe measure of success. Hence, since we are extending that world via low impact to situations where the AI exists, the difference between $A$ and our true goals is not an issue.
}. %
To make this safe, we would want $P(A|\neg X)$ to not be too low -- we have a non-insignificant chance of getting rich even if the AI was actually not present. So a world with $A$ and $\neg X$ is not particularly unusual; therefore, the world with $X$ that is of low impact compared with it is not particularly unusual either. We might even be able to specify $P(A|\neg X)$ precisely -- maybe there is a selection of a thousand stocks, and the AI recommends one; otherwise a random one is selected. Then $A$ is the announcement that that stock outperformed all the others, making $P(A|\neg X)=1/1000$.

Essentially, the AI operates here as a probability pump, pushing us in the direction we might have been able to achieve anyway, but now increasing its probability. It gives us advice that that a human/WBE/chatbot/random selector could have given us, but gives us the best version of that advice.

\subsection{High impact from multiple conditional low impact}

Take the following hypothetical. Assume that an asteroid is heading to Earth, and that we wish to aim a laser to deflect it. If we ask a single low impact AI to do so, low impact will prevent it from doing so.

But now imagine there are two AIs, Alice and Bob, and that they are activated by events $X$ and $Y$ respectively. Each is indifferent or nearly indifferent to how to world ends up given that the other is turned on. However, conditional on Bob being turned off, Alice is low impact but desires to output the correct $x$-coordinate for aiming the laser (this defines $u_{\textrm{Alice}}$). Likewise, conditional on Alice being turned off, Bob is low impact, but desires to output the correct $y$-coordinate.

So, because Alice only cares about worlds in which $\neg Y$, even if she's highly confident that $Y$, she'll still output the correct $x$-coordinate. Furthermore, since in the $\neg Y$-worlds, the laser will miss the asteroid for want of a correct $y$-coordinate, she can still count as \emph{conditionally} low impact. And the same, \emph{mutatis mutandis}, goes for Bob. 

\begin{figure}[h!tb]
	\centering
	\includegraphics[width=0.7\textwidth]{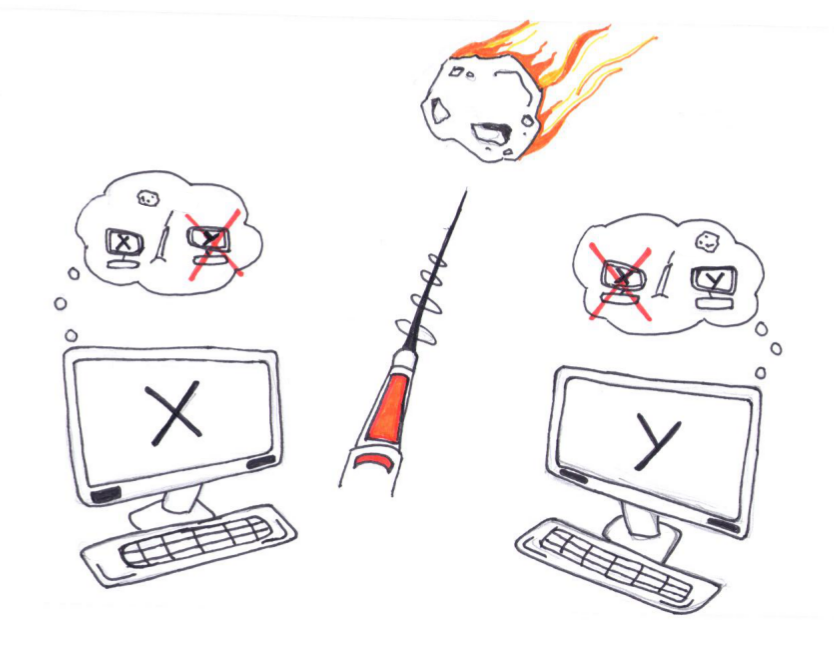}
	\caption{Two AIs, each acting on the assumption that the other one is not active, independently compute their half of the aiming coordinates.}
	\label{comps_get_asteroid}
\end{figure}

So, both Alice and Bob will output the correct coordinates, and we get a large impact -- as long as we set the probabilities of $\neg X$ and $\neg Y$ low enough. But all negative AI impacts -- such as sneaking something dangerous into the coordinates, or escaping and optimising the world -- can still be precluded. This is illustrated by figure \ref{comps_get_asteroid}.

\begin{figure}[h!tb]
	\centering
	\includegraphics[width=0.7\textwidth]{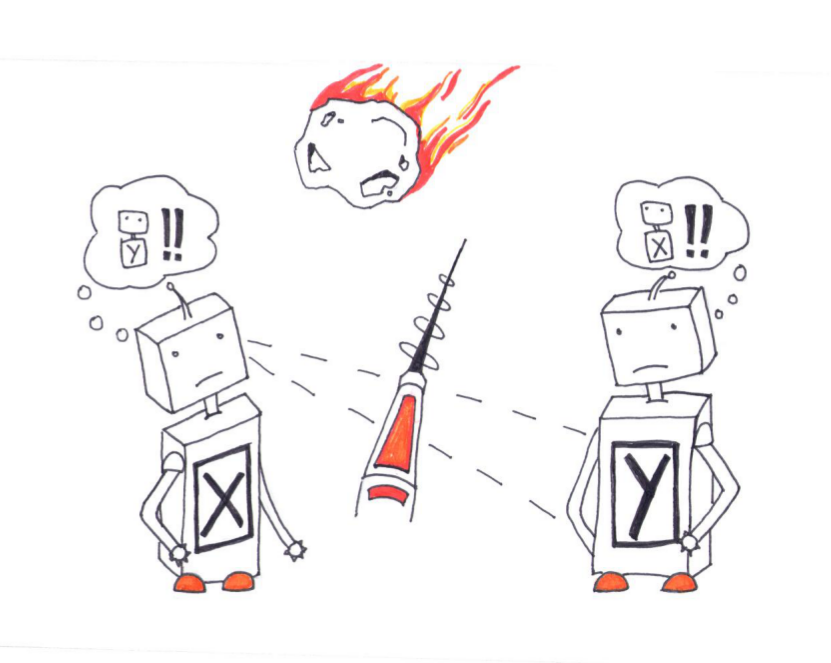}
	\caption{Two robots, each acting on the assumption that the other one is not active, notice the presence of the other and their behaviour becomes undefined.}
	\label{robots_miss_asteroid}
\end{figure}

This doesn not work if the two AIs are robots that can move around and observe that the other is, in fact, active. In that case, they have to act as if the other is not active while observing that they are, thus causing them to come up with complicated explanations as to why this is so. Their behaviour is thus likely to become undefined. In particular, they are probably not going to act as low impact laser aimers, see figure \ref{robots_miss_asteroid}.

\subsection{Extension of category concepts}

The previous `aiming laser' example does not allow the AIs to have a (desired and targeted) high impact in general situations. It does not, for instance, allow the AIs to walk around and notice that the other AI is actually turned on -- it only works because they don't know the other AI was activated. 

How might we make this approach more broadly applicable? One potential idea is for the AI to derive general concepts of low impact conditional on $\neg X$ -- concepts such as `don't fill the universe with self-replicating probes' or `don't take over the world' conditional on the other AI not being activated. Then we could hope that these concepts can be  generalised to cover the $X$ situation as well. Specifically, we might imagine that the AI creates the code for a subagent, using these concepts, and that the subagent goes out and does things in the world, maybe interacting with the other subagents, and physically aiming the laser.

There are two fundamental challenges here. The first is that we don't want the subagent to generalise low impact to include `not outputting the correct $y$-coordinates'. This could be avoided either by requiring that the categories used be roughly independent of $X$ vs $\neg X$, or by using some of the methods of section \ref{res:con} (a successful deflection of the asteroid seems a pretty clear outcome that could be conditioned on).

The second issue is more problematic. Consider the concept $C$: ``if $\neg X$, nothing happens, if $X$, a nuclear war starts''. This concept can describe low impact, conditional on $\neg X$, but is obviously terrible at extending low impact to the $X$ case.

Now $C$ looks disjunctive and artificial, and we'd like to rule concepts like this out. But it turns out to be hard; there are no easy ways to distinguish unnatural disjunctive categories from natural ones (see the issues with Grue and Bleen versus Blue and Green for a very relevant example of this kind of problem \cite{grue_bleen}). Research in this possibility is ongoing\footnote{%
See the corresponding author's work at \url{http://lesswrong.com/lw/mbq/the_president_didnt_die_failures_at_extending_ai/} , \url{http://lesswrong.com/lw/mbp/green_emeralds_grue_diamonds/} , \url{http://lesswrong.com/r/discussion/lw/mbr/grue_bleen_and_natural_categories/} , and \url{http://lesswrong.com/r/discussion/lw/mfq/presidents_asteroids_natural_categories_and/}.
}.

\section{Known issues}

There are a couple of difficulties with the whole low impact approach. The general one, common to the Friendly AI approach as well, is that the AI may think of a loophole that we cannot; the risk of this is low the more analysis we do and the better we understand the situation.

But there are more specific issues. The $R$ in equation (\ref{penalty:equation}) is not a utility function; instead it is a penalty function that the AI itself calculates, using its own probability modules $P'$ (and in one case it uses this to estimate the output of an idealised probability module $P$ -- see Section \ref{undetect}).

What would happen if the AI self-modifies and changes $P'$? There is a meta-argument that this shouldn't matter -- the AI is committed to low impact, and therefore it will ensure that its future copies also have at least as much low impact. This argument does not feel fully reassuring, however, and it is very possible that some bad programming would be disastrous. For instance, we want $P'$ to be properly abstractly defined, not labeled as (the equivalent of) `the output of that box over there', as `that box over there' can always be modified physically. But it might not always be clear how the agent is formally defining $P'$; this is especially the case if there is some implicit probability estimate happening elsewhere in the AI. For instance, what if the pre-processing of inputs to $P'$ was very important, and $R$ was defined sloppily enough that changing the pre-processing could change its definition?

The more general issue is that any goal that is not a utility function is unstable \cite{armstrong_Omohundro_2008}, in that an agent with one will seek to change it if they can\footnote{%
Utility functions are generally seen as stable, but even there there are subtleties. Because utility functions form a kind of affine space, \emph{any} utility function being unstable means \emph{almost all} of them are. To see why, note that a stable utility function mixed with or added to an unstable one will be unstable. It still remains the case, though, that nearly all utility functions we could naturally think of, are stable.
}.

The corresponding author intends to analyse the issue in subsequent papers: what do unstable goals tend to if the agent can self-modify? This would both be useful to preserve the needed parts of unstable goals (such as the low impact) and might also allow us to express things like low impact in a clear, and we hope instructive, utility function format.

There is also the risk that a series of low impact AIs, through their individual decisions, end up having a large impact even if no specific AI does so. That particular problem can be addressed by making the AIs indifferent to the existence/outputs of the other AIs\footnote{%
See the corresponding author's post at \url{http://lesswrong.com/r/discussion/lw/lyh/utility_vs_probability_idea_synthesis/}
}.
However, this is a patch for a particular issue, rather than a principled declaration that there are no further issues. Such a declaration or proof would be of great use, as repeated patching of an idea does not end when the idea is safe, but when we can no longer think of reasons it is unsafe.

\bibliographystyle{alpha}
\bibliography{../ref}

\end{document}